\documentclass[sigconf,natbib=true]{acmart}

\AtBeginDocument{}

\copyrightyear{2026}
\acmYear{2026}
\setcopyright{cc}
\setcctype{by}
\acmDOI{XXXXXXX.XXXXXXX}
\acmISBN{979-8-4007-2539-5/2026/11}

\acmConference[KEIR@CIKM 2026]{The 3rd Workshop on Knowledge-Enhanced Information Retrieval at Conference on Information and Knowledge Management}{November 8, 2026}{Rome, Italy}

\usepackage{lipsum}
\usepackage{graphicx}
\usepackage{multirow}
\usepackage{tabularx}
\usepackage{booktabs}
\usepackage{makecell}
\usepackage{array}
\usepackage{float}
\usepackage{placeins}
\usepackage{bm}

\newcommand{\ours}{MiroGraph}
\usepackage{xcolor}
\newif\ifrevision
\revisionfalse
\ifrevision
  \newcommand{\rev}[1]{\textcolor{blue}{#1}}
\else
  \newcommand{\rev}[1]{#1}
\fi

\begin{document}

\title{Bridging Concepts: Multi-Agent Reasoning over Scientific Concept Graphs for High-Impact Research Prediction}

\author{Donggyu Yoon}
\authornote{These authors contributed equally to this work.}
\affiliation{\institution{School of Computing, KAIST}\city{Daejeon}\country{Republic of Korea}}
\email{justinyoonwk@kaist.ac.kr}

\author{Minsung Hwang}
\authornotemark[1]
\affiliation{\institution{School of Computing, KAIST}\city{Daejeon}\country{Republic of Korea}}
\email{hminsung@kaist.ac.kr}

\author{Juneha Baek}
\affiliation{\institution{College of Business, KAIST}\city{Seoul}\country{Republic of Korea}}
\email{8666baek@kaist.ac.kr}

\author{Donghyuk Shin}
\affiliation{\institution{College of Business, KAIST}\city{Seoul}\country{Republic of Korea}}
\email{dhs@kaist.ac.kr}

\author{Joyce Jiyoung Whang}
\authornote{Corresponding author}
\affiliation{\institution{Department of AI Computing, KAIST}\city{Daejeon}\country{Republic of Korea}}
\email{jjwhang@kaist.ac.kr}

\begin{abstract}
Within a research community, new topics often emerge when previously
unconnected concepts are first studied together. We construct
\emph{scientific concept graphs} (SCGs)---domain knowledge graphs whose
nodes are fine-grained research concepts extracted from paper text by
LLMs and whose edges capture their co-occurrence in papers---and study
the task of forecasting high-impact concept pairs, i.e., discovering
concept pairs that are currently unconnected but will have high impact
when connected. To this end, we introduce new benchmark datasets
constructed from CIKM, KDD, and SIGIR conference papers, with varied
prediction settings, including impact forecasting for emerging concepts
unseen during training. We propose \mbox{\ours}, a
\textbf{\underline{M}}ult\textbf{\underline{i}}-agent
\textbf{\underline{r}}eas\textbf{\underline{o}}ning framework with
scientific concept \textbf{\underline{Graph}}s, in which a feature
designer agent designs predictive features for a given concept pair from
structural, semantic, and trend perspectives, and then three LLM agents
compute the actual feature values from the enclosing subgraph of the
concept pair. Using these feature values, a classifier designer agent
constructs a classifier that predicts the final impact labels. \ours\
outperforms various impact forecasting and temporal link prediction
methods.
\end{abstract}



\maketitle

\section{Introduction}
\label{sec:introduction}

Influential research often arises from combinations of distinct research concepts, and thus bridging concepts with high-impact potential can provide valuable insight into future research directions~\cite{combinations}. Recent work has explored a task of forecasting high-impact concept pairs on scientific concept graphs (SCGs), where research concepts are represented as nodes, and their co-occurrence in papers is captured by edges~\cite{impact4cast}. However, Impact4Cast~\cite{impact4cast}, the most representative work, depends on keyword-level concept extraction and manual processing for constructing SCGs, limiting its scalability and the granularity of extracted concepts. It also applies a simple neural network to 141 handcrafted features derived from SCGs to forecast high-impact concept pairs, restricting the model to predefined metrics.

\begin{figure*}[ht!]
\centering
\includegraphics[width=\textwidth]{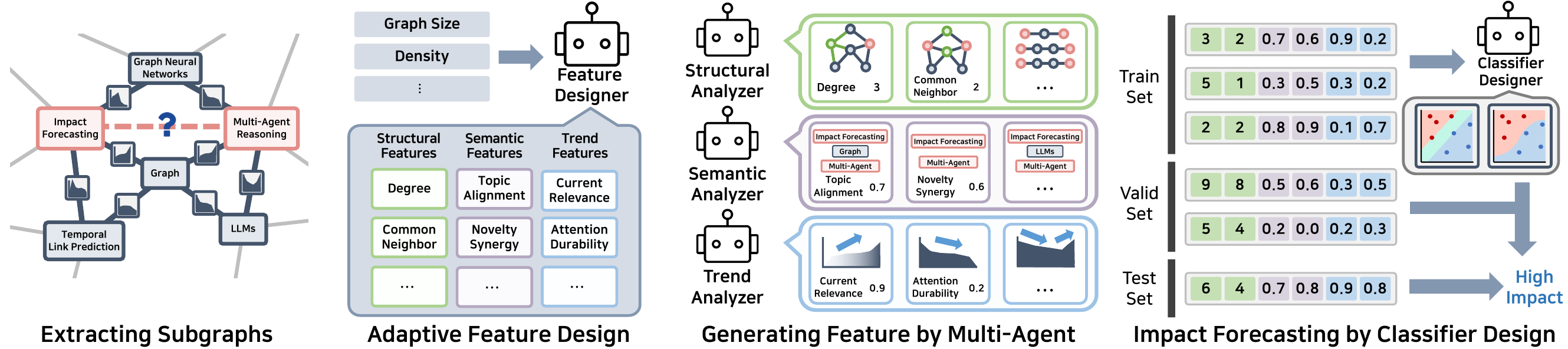}
\caption{Overview of \ours. \ours\ extracts an enclosing subgraph and employs a feature designer agent to determine predictive feature types based on statistics of the SCG. Three LLM agents compute feature values from structural, semantic, and trend perspectives. A classifier designer agent constructs classifiers to predict the final impact label based on the feature values.}
\Description{
The figure illustrates the overall pipeline of MiroGraph. On the left, an enclosing subgraph is extracted for a candidate concept pair from a scientific concept graph. A feature designer agent then receives summary statistics of the entire SCG and determines structural, semantic, and trend feature types for the target concept pair. In the middle, three LLM-based agents, a Structural Analyzer, a Semantic Analyzer, and a Trend Analyzer, generate corresponding feature values from the enclosing subgraph. The generated features are normalized and combined into a unified feature representation. On the right, a classifier designer generates and selects classifiers using the generated feature representations from the training and validation sets to predict whether the concept pair is high-impact.
}
\label{fig:fig2}
\end{figure*}

In this paper, we propose \ours, which automates the entire process
from feature design to classifier construction: a feature designer agent
produces a predictive feature schema from structural, semantic, and trend
perspectives, three LLM agents compute the corresponding values over an
enclosing subgraph, and a classifier designer agent builds the final
classifier. We evaluate \ours\ on three SCGs built from major information
retrieval and data mining conferences, where it achieves the best
impact forecasting performance in all settings. Ablations show that the automated feature and classifier design is essential, especially when generalizing to concepts unseen during training. Our SCG datasets and code are available
at \url{https://github.com/bdi-lab/MiroGraph}.

\section{Related Work}
Prior work has explored predicting the impact of individual papers on citation networks~\cite{cody, cines, hotness_prediction_bkg, tgat_topic_trend_prediction, award_prediction}. More recently, several studies have explored forecasting whether pairs of research concepts will be jointly studied using SCGs~\cite{science4cast, quantum_emergent_connections, interdisciplinary_link_prediction, m1, m2, m3, m4, m5, m6, m8, augur}. However, they do not incorporate citation counts, and thus cannot predict whether such co-occurrences will lead to high-impact research. To forecast high-impact concept pairs on SCGs, Impact4Cast~\cite{impact4cast} applies simple neural networks on handcrafted features. LLMs have also been used to automatically engineer features for tabular prediction~\cite{caafe, featllm}, where features are derived from predefined attributes of each instance. In contrast, signals for a concept pair in SCGs are dispersed across its relational and temporal context; \ours\ adapts the feature schema to graph-level statistics, computes structural, semantic, and trend features from each enclosing subgraph with specialized agents, and automates classifier construction.

Link prediction in dynamic graphs addresses the task of predicting future edge formation from historical interactions ~\cite{tgat,dygformer,cawn,graphmixer,tgn,jodie,dyrep}. While these methods can be adapted to our task, they cannot consider the semantic meanings of concepts, which \ours\ captures together with structural and trend signals.

\section{Construction of SCG Datasets}
\label{sec:dataset}

We develop an automated pipeline for constructing SCGs with fine-grained research concepts using hierarchical concept extraction and LLM-based concept canonicalization. An SCG $G = (\mathcal{V}, \mathcal{E})$ is built from a set of papers $\mathcal{P}$, where each node $v \in \mathcal{V}$ represents a concept and an edge $(u,v) \in \mathcal{E}$ indicates that the two concepts have been jointly studied in a paper. Each edge is associated with yearly citation counts as attributes; to prevent temporal leakage, each snapshot graph only includes citations received up to its snapshot year. We collect titles and abstracts of all papers available in the online proceedings of CIKM (1993--2025), KDD (1995--2025), and SIGIR (1978--2025) and use GPT-4o-mini~\cite{gpt4omini} to organize each paper into structured descriptions including research objectives, key entities, and domain contexts. Based on the structured descriptions, GPT-4o-mini~\cite{gpt4omini} extracts concepts organized into a five-level hierarchy ranging from broad research areas to fine-grained technical concepts. We retain only concepts from Levels 2--5 to remove overly broad concepts. After filtering out concepts that appear fewer than two times across the corpus, we canonicalize the remaining concepts. Specifically, we apply HDBSCAN~\cite{hdbscan} and assign each concept to all clusters whose membership probability is at least $0.95$ of its maximum value, since a concept can appear in multiple contexts. For each cluster, we prompt GPT-4o-mini~\cite{gpt4omini} to designate a representative concept name. To validate this canonicalization step, we manually inspected randomly sampled clusters and confirmed that the assigned representative names subsume their members. An SCG is constructed by connecting concepts that co-occur in the same paper. The citation counts are retrieved and aggregated from OpenAlex~\cite{openalex}.

\begin{table}[t!]
\centering
\setlength{\tabcolsep}{3.7pt}
\caption{Statistics of our SCG benchmark datasets. $|\mathcal{P}|$ denotes the number of papers used to construct each snapshot graph.}
\label{tab:dataset_statistics}
\begin{tabular}{ccccccc}
\toprule
Venue & Setting & $|\mathcal{V}|$ & $|\mathcal{E}|$ & $|\mathcal{P}|$ & $|\mathcal{D}_{\text{train}}|$ & $|\mathcal{D}_{\text{test}}|$ \\
\midrule
\multirow{2}{*}{\textsc{SCG-CIKM}}
& Test & 6,544 & 74,887 & 2,254 & 1,916 & 338 \\
& Eval & 6,059 & 61,862 & 1,948 & 1,610 & 338 \\
\midrule
\multirow{2}{*}{\textsc{SCG-KDD}}
& Test & 3,833 & 41,685 & 3,196 & 2,716 & 480 \\
& Eval & 3,529 & 34,561 & 2,490 & 2,010 & 480 \\
\midrule
\multirow{2}{*}{\textsc{SCG-SIGIR}}
& Test & 3,188 & 34,445 & 2,734 & 2,324 & 410 \\
& Eval & 2,925 & 27,465 & 2,618 & 2,208 & 410 \\
\bottomrule
\end{tabular}
\end{table}

Given a snapshot of SCGs constructed from publications up to a certain year, the task is to predict whether a currently unconnected concept pair will later form a high-impact connection. Following Impact4Cast~\cite{impact4cast}, a concept pair is labeled as positive if it newly appears in papers during a target period and the accumulated citations of those papers reach or exceed a predefined threshold $IR$ (Impact Range). Concept pairs that newly appear in the future window but whose accumulated citations remain below $IR$ are treated as hard negatives. We sample additional negatives from unconnected concept pairs that are not positive pairs, using a mixed strategy that prioritizes hard negatives among concepts appearing in positive pairs and fills the remaining samples from all other unconnected concept pairs. The resulting datasets have a 1:1 positive-to-negative ratio, and hard negatives account for 36.1\% of the negatives. We set $IR=5$ following Impact4Cast~\cite{impact4cast} for comparability, such that positive pairs represent approximately the top 15\% of newly formed concept pairs based on accumulated citations.

\begin{table*}[t!]
\centering
\setlength{\tabcolsep}{3.5pt}
\caption{Main results in Macro F1 ($\uparrow$). Best results are bolded, and second-best results are underlined. An asterisk (*) indicates a statistically significant improvement over the best baseline based on a paired approximate randomization test ($p < 0.05$).}
\label{tab:main_results}
\begin{tabular}{lcccccc}
\toprule
Method 
& \multicolumn{2}{c}{\textsc{SCG-CIKM}} 
& \multicolumn{2}{c}{\textsc{SCG-KDD}} 
& \multicolumn{2}{c}{\textsc{SCG-SIGIR}} \\
\cmidrule(lr){2-3}\cmidrule(lr){4-5}\cmidrule(lr){6-7}
& Test & Eval & Test & Eval & Test & Eval \\
\midrule
Impact4Cast~\cite{impact4cast}
& 0.6633$\pm$0.0197
& 0.5865$\pm$0.0106
& \underline{0.6030$\pm$0.0065}
& 0.5811$\pm$0.0074
& \underline{0.6143$\pm$0.0198}
& 0.6106$\pm$0.0063 \\
TGAT~\cite{tgat}
& 0.6439$\pm$0.0062
& 0.5465$\pm$0.0069
& 0.5645$\pm$0.0025
& 0.5662$\pm$0.0013
& 0.5627$\pm$0.0061
& 0.5527$\pm$0.0148 \\
DyGFormer~\cite{dygformer}
& 0.6785$\pm$0.0173
& \underline{0.6264$\pm$0.0178}
& 0.5764$\pm$0.0064
& 0.5736$\pm$0.0081
& 0.6065$\pm$0.0019
& 0.5951$\pm$0.0076 \\
CAWN~\cite{cawn}
& 0.6311$\pm$0.0086
& 0.5699$\pm$0.0049
& 0.5642$\pm$0.0024
& 0.5727$\pm$0.0018
& 0.5747$\pm$0.0180
& 0.6001$\pm$0.0098 \\
GraphMixer~\cite{graphmixer}
& 0.5844$\pm$0.0938
& 0.5162$\pm$0.0227
& 0.5724$\pm$0.0073
& 0.5194$\pm$0.0502
& 0.5566$\pm$0.0582
& \underline{0.6118$\pm$0.0131} \\
TGN~\cite{tgn}
& \underline{0.7292$\pm$0.0090}
& 0.5312$\pm$0.0111
& 0.5319$\pm$0.0256
& 0.5052$\pm$0.0114
& 0.5344$\pm$0.0169
& 0.5649$\pm$0.0182 \\
JODIE~\cite{jodie}
& 0.5966$\pm$0.0123
& 0.5272$\pm$0.0604
& 0.5256$\pm$0.0152
& 0.4888$\pm$0.0075
& 0.5423$\pm$0.0260
& 0.5178$\pm$0.0066 \\
DyRep~\cite{dyrep}
& 0.5735$\pm$0.0142
& 0.5285$\pm$0.0155
& 0.5236$\pm$0.0344
& 0.4877$\pm$0.0096
& 0.5545$\pm$0.0149
& 0.4983$\pm$0.0148 \\
\midrule
GPT-4o-mini
& 0.5947$\pm$0.0011
& 0.5976$\pm$0.0032
& 0.5761$\pm$0.0050
& 0.5780$\pm$0.0013
& 0.5930$\pm$0.0068
& 0.5995$\pm$0.0013 \\
GPT-4o
& 0.5829$\pm$0.0047
& 0.5771$\pm$0.0076
& 0.5997$\pm$0.0034
& \underline{0.6009$\pm$0.0014}
& 0.5751$\pm$0.0079
& 0.5758$\pm$0.0062 \\
GPT-4o-mini w/ subgraph
& 0.3693$\pm$0.0059
& 0.3718$\pm$0.0091
& 0.3982$\pm$0.0049
& 0.4063$\pm$0.0077
& 0.4479$\pm$0.0130
& 0.4333$\pm$0.0070 \\
GPT-4o w/ subgraph
& 0.4408$\pm$0.0107
& 0.4198$\pm$0.0107
& 0.4528$\pm$0.0102
& 0.4455$\pm$0.0224
& 0.4901$\pm$0.0172
& 0.4745$\pm$0.0086 \\
\midrule
\textbf{\ours}
& \textbf{0.7501*$\pm$0.0217}
& \textbf{0.6435*$\pm$0.0032}
& \textbf{0.6878*$\pm$0.0222}
& \textbf{0.6334*$\pm$0.0051}
& \textbf{0.6818*$\pm$0.0096}
& \textbf{0.6157$\pm$0.0209} \\
\bottomrule
\end{tabular}
\end{table*}

We consider two prediction settings: \emph{Test} and \emph{Eval}. In the \emph{Test} setting, we construct a snapshot graph $G_\text{test}$ using papers published up to 2023 and define the target period as 2024–2025. Candidate concept pairs emerging during 2024–2025 are split into a training set $\mathcal{D}_{\text{train}}$ and a held-out test set $\mathcal{D}_{\text{test}}$ with an $85{:}15$ ratio. In the \emph{Eval} setting, the prediction target remains identical to the \emph{Test} setting, and thus the same $\mathcal{D}_{\text{test}}$ is used for evaluation. We additionally construct a snapshot graph $G_\text{train}$ using papers published up to 2021 and define the target period as 2022–2023, from which the training set for the \emph{Eval} setting is derived. Since the model is trained on earlier graph snapshots and evaluated on later emerging concepts, the \emph{Eval} setting naturally introduces unseen concepts: 51.00\%, 28.00\%, and 41.96\% of the concepts in $\mathcal{D}_{\text{test}}$ do not appear in its training set on \textsc{SCG-CIKM}, \textsc{SCG-KDD}, and \textsc{SCG-SIGIR}, respectively. We place particular emphasis on \emph{Eval}, as it additionally tests temporal generalization to concepts unseen during training. Table~\ref{tab:dataset_statistics} summarizes the resulting datasets. Implementation details, prompts, and citation distributions are provided in the code.

\begin{table*}[t!]
\centering
\setlength{\tabcolsep}{4.5pt}
\caption{Representative features and corresponding values for the pair \textit{Knowledge Tracing Methods} and \textit{Heterogeneous Graph}
}
\label{tab:case_study_agents}
\begin{tabular}{llcl}
\toprule
\textbf{Perspective} & \textbf{Feature} & \textbf{Value} & \textbf{Interpretation} \\
\midrule
\multirow{2}{*}{Structural}
& Shortest path length & $4$
& The pair is connected only through distant multi-hop bridges. \\
& Common neighbors & $0.00$
& The pair is not a trivial local-closure case. \\
\midrule
\multirow{2}{*}{Semantic}
& Novelty synergy & $0.75$
& The combination suggests a novel cross-area integration. \\
& Cross-domain fit & $0.70$
& The pair bridges educational modeling and graph representation learning. \\
\midrule
\multirow{2}{*}{Trend}
& Current relevance & $0.75$
& Both concepts are relevant to recent research directions. \\
& Attention durability & $0.73$
& The topic is expected to sustain attention beyond a short-lived spike. \\
\bottomrule
\end{tabular}
\end{table*}

\section{\ours: Forecasting High-Impact Research Topics by Multi-Agent Reasoning}
\label{sec:models}
Figure~\ref{fig:fig2} illustrates the overview of \ours. To predict the impact of a query concept pair $(u,v)$, we first extract an enclosing subgraph that provides localized relational context from the global SCG. Specifically, we collect all \emph{induced paths} connecting $u$ and $v$ whose length is at most four, and construct an induced subgraph over the nodes appearing in these induced paths. The resulting subgraph is serialized into a JSON-formatted edge list. Throughout the framework, agents that never observe concept
identities---the feature designer, the structural analyzer, and the
classifier designer---cannot leak pretrained knowledge, while the
semantic and trend analyzers, which do observe concepts, use
GPT-4o~\cite{gpt4o}, whose knowledge cutoff precedes the target period.

\subsection{Adaptive Feature Designing}
\label{sec:adaptive_feat_design}
We employ a feature designer agent implemented with GPT-5.4~\cite{gpt54} to design predictive feature schemas that may be useful for predicting whether the target concept pair will form a high-impact connection. Since SCGs can exhibit different structural characteristics, the agent is additionally provided with summary statistics of the entire SCG, such as its size, density, and degree distribution, enabling it to adapt the feature schema to the characteristics of the input graph.  The resulting feature schema consists of three perspectives: \emph{structural}, \emph{semantic}, and \emph{trend}. Structural features capture topological information such as connectivity patterns of intermediate concepts linking the target pair. Semantic features characterize how the research themes of concepts in the subgraph relate to each other. Trend features capture citation-based temporal signals, such as yearly citation dynamics among related concepts.

\subsection{Multi-Agent Feature Computing}
\label{sec:multi_perspective_scoring}
Given the enclosing subgraph for a query pair, we employ three LLM agents to compute feature values. The \emph{Structural Analyzer}, implemented with GPT-5.4~\cite{gpt54}, generates executable code whose execution yields the structural feature values. The \emph{Semantic Analyzer} and the \emph{Trend Analyzer} compute semantic and trend feature values by directly analyzing the enclosing subgraph, using GPT-4o~\cite{gpt4o} as discussed above. The resulting values are normalized and combined into the final feature representation.

\subsection{Impact Forecasting by Classifier Design}
\label{sec:classifier_generation}

To predict the impact label from the feature values, rather than fixing a specific classifier architecture, we introduce a \emph{Classifier Designer} agent implemented with GPT-5.4~\cite{gpt54}, which designs candidate classification models using the feature values derived from the training set $\mathcal{D}_{\text{train}}$. We split \(\mathcal{D}_{\text{train}}\) into training and validation subsets for model learning and model selection, respectively, at a $70{:}15$ ratio (i.e., about 82\% and 18\% of $\mathcal{D}_{\text{train}}$). In the \emph{Test} setting, this corresponds to a $70{:}15{:}15$ train/validation/test split of all candidate pairs. The agent produces multiple classifiers that map the feature values to the final high-impact label, and the model achieving the best performance on the validation subset is selected as the final classifier. Since the feature designer and classifier designer are invoked once per dataset, only the analyzers are called per candidate pair. For reproducibility, the prompts, generated feature schemas, feature values, and classifier code are available in our repository.

\begin{figure}[t]
\centering
\includegraphics[width=\linewidth]{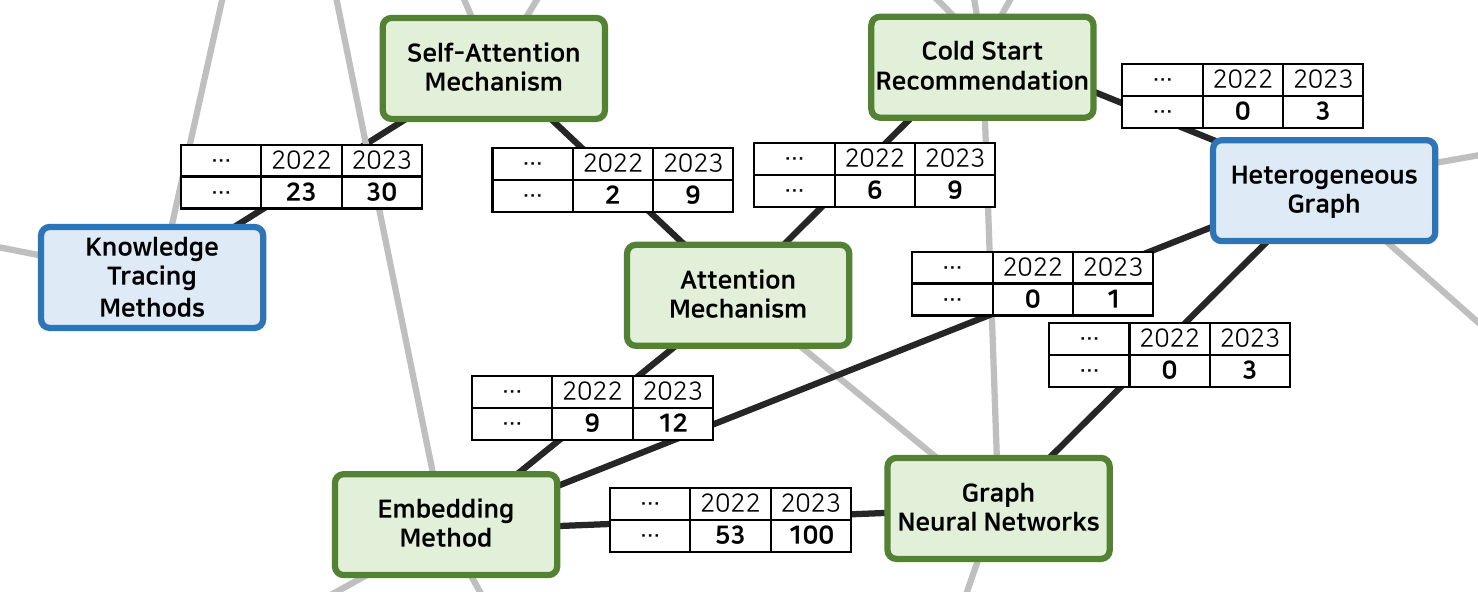}
\caption{Enclosing subgraph extracted for `\textit{Knowledge Tracing Methods}' and `\textit{Heterogeneous Graph}'. For clarity, only a subset of nodes, edges, and edge attributes is visualized.}
\Description{The figure shows an enclosing subgraph extracted for a concept pair `Knowledge Tracing Methods' and `Heterogeneous Graph.' The two target concepts are connected through several intermediate research concepts, including Self-attention mechanism, Attention mechanism, Embedding Method, Graph neural networks, and Cold start recommendation. Edges are annotated with citation counts aggregated for recent years, illustrating the temporal changes in interactions among related concepts.}
\label{fig:case_study_agents}
\end{figure}

\begin{table}[t!]
\centering
\setlength{\tabcolsep}{6pt}
\caption{Ablation study on \textsc{SCG-CIKM}. Best results are bolded.}
\label{tab:ablation}
\begin{tabular}{lcc}
\toprule
Method 
& \textsc{SCG-CIKM}(\textit{Test}) 
& \textsc{SCG-CIKM}(\textit{Eval}) \\
\midrule
\ours 
& $\bm{0.7501 \pm 0.0217}$ 
& $\bm{0.6435 \pm 0.0032}$ \\
w/o structure 
& $0.5844 \pm 0.0041$ 
& $0.5805 \pm 0.0133$ \\
w/o semantic 
& $0.7297 \pm 0.0089$ 
& $0.6192 \pm 0.0196$ \\
w/o trend 
& $0.7373 \pm 0.0093$ 
& $0.6065 \pm 0.0010$ \\
single score 
& $0.6939 \pm 0.0050$ 
& $0.5355 \pm 0.0041$ \\
MLP 
& $0.6928 \pm 0.0204$ 
& $0.5731 \pm 0.0195$ \\
\bottomrule
\end{tabular}
\end{table}

\section{Experiments}
\label{sec:experiments}

We compare \ours\ with Impact4Cast \cite{impact4cast}, the most relevant baseline. We also adapt temporal link prediction models, including DyRep~\cite{dyrep}, JODIE~\cite{jodie}, TGN~\cite{tgn}, TGAT~\cite{tgat}, CAWN~\cite{cawn}, DyGFormer~\cite{dygformer}, and GraphMixer~\cite{graphmixer}, which handle nodes not observed during training. For temporal link prediction baselines, we treat each pair as an edge only if the accumulated citations within two years reach or exceed $IR$. We further consider LLM baselines implemented with GPT-4o~\cite{gpt4o} and GPT-4o-mini~\cite{gpt4omini}, which directly predict the impact label from the concept pair. We also compare variants that take the enclosing subgraph as a direct input prompt. We report the mean and standard deviation over three runs.

\subsection{Experimental Results}
\label{sec:results}

Table~\ref{tab:main_results} shows that \ours\ achieves the best performance across all settings in Macro F1, with statistically significant gains in five of the six settings. \rev{Baselines that perform well in the \emph{Test} setting can degrade substantially in the \emph{Eval} setting, which requires generalization to unseen concepts; e.g., TGN drops from 0.7292 to 0.5312 on \textsc{SCG-CIKM}. In contrast, \ours\ remains the best in both settings on all datasets.} Interestingly, directly providing the subgraph to an LLM leads to lower Macro F1 than using the same LLM without subgraph information. This suggests that the gains of \ours\ stem from the proposed multi-agent framework rather than from the LLMs’ internal knowledge or direct reasoning over the subgraph.

\subsection{Case Study and Ablation Study}
\label{sec:case_study}
Figure~\ref{fig:case_study_agents} visualizes the enclosing subgraph for the concept pair `\textit{Knowledge Tracing Methods}' and `\textit{Heterogeneous Graph}' under the \emph{Test} setting of \textsc{SCG-CIKM}. Table~\ref{tab:case_study_agents} reports representative features determined by the feature designer, along with the feature values generated by each analyzer and their interpretations obtained through additional prompting. This case illustrates the kinds of signals captured by each perspective, although such post-hoc interpretations do not by themselves establish the basis of the prediction.

Table~\ref{tab:ablation} reports ablations on \textsc{SCG-CIKM} with variants that exclude one of the three analysis agents (\textbf{w/o structure}, \textbf{semantic}, \textbf{trend}), remove the feature designer agent and instead let each analyzer generate only a single score (\textbf{single score}), and replace the classifier designer agent with a standard MLP classifier (\textbf{MLP}). All variants underperform \ours, confirming the role of each component. Notably, the performance degradation of \textbf{single score} and \textbf{MLP} is substantially larger in the \emph{Eval} setting than in the \emph{Test} setting, indicating that automated feature design and classifier construction play an important role in generalization under the more challenging settings.

\section{Conclusion}
We proposed \ours, a multi-agent reasoning framework in which agents design predictive features for concept pairs over SCGs, compute them from enclosing subgraphs, and construct the final classifier. On our benchmark datasets, \ours\ achieves the best performance in all evaluation settings. As future work, we plan to extend SCGs into richer knowledge graphs
and leverage knowledge graph representation learning and
reasoning~\cite{ingram,hynt,vista,maypl,krepe} to better generalize to
concepts unseen during training.

\begin{acks}
This work was supported by the NRF of Korea grant funded by MSIT (90\% from RS-2025-00559066, 10\% from RS-2022-NR068758).
\end{acks}

\section*{GenAI Usage Disclosure}
In compliance with ACM's Authorship Policy and CIKM submission guidelines, we disclose that generative AI tools were employed as part of this research. Specifically, GPT-4o-mini was utilized to organize paper descriptions, extract research concepts, and designate representative concept names for dataset construction (Section~\ref{sec:dataset}). Within \ours\ (Section~\ref{sec:models}), GPT-5.4 was employed to design feature schemas, generate code for structural features, and design classifiers, while GPT-4o was used to compute semantic and trend feature values. GPT-4o and GPT-4o-mini also served as baselines in our experiments. GenAI tools were used only for grammar and language editing of the manuscript, and all content was written and verified by the authors, who remain fully accountable for the integrity of this work.

\bibliographystyle{ACM-Reference-Format}
\bibliography{cikm_2026_short}

\end{document}